\documentclass[letterpaper, 10 pt, conference]{ieeeconf}  

\IEEEoverridecommandlockouts                              
\title{\LARGE \bf
Consistency issues in Gaussian Mixture Models reduction algorithms
}

\author{Alessandro D'Ortenzio and Costanzo Manes
\thanks{A.\ D'Ortenzio and C.\ Manes are with Department of Information Engineering, Computer Science and Mathematics, University of L'Aquila, 67100 L'Aquila, Italy,
        {\tt\footnotesize alessandro.dortenzio@graduate.univaq.it, 
        costanzo.manes@univaq.it.}}%
\thanks{This work has been supported by the Italian Government under 
CIPE resolution n. 70/2017, Centre of excellence EX-EMERGE}%
}

\usepackage{bbm}
\usepackage{amsmath}
\usepackage{amssymb}
\usepackage{pifont}

\DeclareMathOperator*{\argmin}{arg\,min}
\usepackage{multirow}
\usepackage{graphicx}
\usepackage{epsfig}
\usepackage{color}
\usepackage{dsfont}

\DeclareMathAlphabet\mathbfcal{OMS}{cmsy}{b}{n}

\def\trace{\textit{tr}}
\def\Ncal{\mathcal{N}}
\def\Ncalbold{\mathbfcal{N}}
\def\Nbar{\bar{N}}

\def\Sigmabar{\,\overline{\!\Sigma}}
\def\Sigmahat{\widehat{\Sigma}}
\def\Sigtilde{\widetilde{\Sigma}}
\def\Sigbold{\boldsymbol{\Sigma}}
\def\Sigboldbar{\,\overline{\!\boldsymbol{\Sigma}}}
\def\Thetabar{\,\overline{\!\Theta}}

\def\muhat{\hat{\mu}}
\def\mubold{\boldsymbol{\mu}}
\def\muboldbar{\boldsymbol{\bar{\mu}}}
\def\what{\hat{w}}
\def\wbar{\bar{w}}
\def\wtilde{\tilde{w}}
\def\mubar{\bar{\mu}}
\def\mutilde{\tilde{\mu}}
\def\ones{ \mathds{1} }
\def\wbold{\boldsymbol{w}}
\def\wboldtilde{\boldsymbol{\tilde w}}
\def\wboldbar{\boldsymbol{\bar w}}
\def\DKL{D_{K\!L}}
\def\DISE{D_{I\!S\!E}}
\def\DNISE{D_{N\!I\!S\!E}}

\def\KLD{{K\!L\!D}}
\def\ISE{{I\!S\!E}}
\def\NISE{{N\!I\!S\!E}}

\def\Real{\mathbb{R}}

\begin{document}

\maketitle
\thispagestyle{empty}
\pagestyle{empty}

\begin{abstract}
In many contexts Gaussian Mixtures (GM) are used to approximate probability distributions, possibly time-varying.
In some applications the number of GM components exponentially increases over time, and reduction procedures are required to keep them reasonably limited. 
The GM reduction (GMR) problem can be formulated by choosing different measures of the dissimilarity of GMs before and after reduction, like the Kullback-Leibler Divergence (KLD) and the Integral Squared Error (ISE).
Since in no case the solution is obtained in closed form,
many approximate GMR algorithms have been proposed in the past three decades, although none of them provides optimality guarantees. 
In this work we discuss the importance of the choice of the dissimilarity measure 
and the issue of consistency of all steps of a reduction algorithm with the chosen measure.
Indeed, most of the existing GMR algorithms are composed by several steps which are not consistent with a unique measure, 
and for this reason may produce reduced GMs far from optimality. 
In particular, the use of the KLD, of the ISE and normalized ISE is discussed and compared in this perspective.
\end{abstract}

\section{INTRODUCTION}

Gaussian Mixtures (GMs) are a powerful tool often used to approximate probability distributions in very different application areas. 
In some contexts, such as in filtering and estimation, GMs approximate time-varying distributions that must be updated in real-time, and very often the number of components of the GM increase with time.
For instance, in the field of target tracking in clutter \cite{PMBM,GSF,GMPHD}, the Bayesian recursion produces a number of components that exponentially grows with time. 
In cases like this, if no action is taken, the problem becomes computationally intractable after few steps.
To address such a drawback, many algorithms have been proposed in the literature in the last three decades that aimed at reducing the components of a given GM, while maintaining a substantial similarity with the original one.
Most algorithms, such as those in 
\cite{AssaPlataniotis,Crouse,Petrucci,Runnalls,Salmond,GMRC,West,WilliamsTh03,Williams06,DPHEM,CTDGMRA},
combine {\it greedy} coarse reduction steps with refinement steps, all with the common underlying principle of minimizing a given dissimilarity measure (D-measure, for short) between the \textit{original} mixture and its \textit{reduced} version. 
In general, the problem of Gaussian Mixture Reduction (GMR) can be cast into a nonlinear constrained optimization problem, where the optimization variables are the parameters of the reduced GM 
(weights, means and covariances). 
The underlying idea is that most of the information contained in the original GM should be conveyed in the reduced GM (minimal loss).
However, a lot of issues arise when tackling the GMR problem. 
First of all, although there are many D-measures that can be used, none of them has theoretical or practical features that make it preferable to others in all applications.
A common problem shared by all D-measures is the presence of a large number of local minima in the GMR optimization problem.
Some of the most used D-measures are the \textit{Kullback-Leibler Divergence (KLD)} \cite{KLD}, the \textit{Integral Squared Error (ISE)} \cite{Scott} and its normalized version (\textit{NISE}), and the \textit{Squared 2-Wasserstein} distance \cite{AssaPlataniotis,Delon}.
Among these, only the ISE and NISE have closed forms when applied to GMs,
while only the KLD allows a closed form computation of the 
{\it barycenter} of a mixture, which is an important quantity in most reduction procedures.  
 Considering that each D-measure has its own pros and cons, most existing reduction algorithms mix up heterogeneous actions inspired by different D-measures, driven mainly by the ease of computation and weakly supported by theoretical considerations.
 
 In this work the use of \textit{KLD}, \textit{ISE} and \textit{NISE} in GMR algorithms is analyzed and discussed.
 The features of these D-measures are discussed and tested on simple but informative examples, and 
some light is shed on the issue of \textit{consistency} of all steps of a GM reduction algorithm
with a single D-measure. 
Indeed, inconsistent choices, often made in practice, may lead to solutions quite far from optimality. 
The notion of Best Single Gaussian Approximation of a submixture is also introduced and compared with the concept of Barycenter of a mixture.
 
 The paper is organized as follows: in Sec.\ II the GMR problem is formulated; Sec.\ III introduces 
 some D-measures; Sec.\ IV presents the basic steps of a GMR algorithm; 
 Sec.\ V introduces the concepts of BSGA and of Barycenter of a GM; 
 Sec.\ VI summarizes the features of KLD, ISE and NISE;
 Sec. VII reports numerical tests. Conclusions follow.
 
 \smallskip
 \noindent
 {\it Notations.} In this paper $\Real^N_+$ denotes the set of nonnegative vectors in $\Real^N$, 
 the symbol $\ones_N$ denotes a vector of ones in $\Real^N$, and will be used to represent summations in compact form: given $\wbold\in\Real^N$, $\wbold^T \ones_N=\sum_{i=1}^N w_i$.
 The standard simplex $\Delta^{N-1}\subset\Real^N_+$ is defined by all $\wbold\in\Real^N_+$: $\wbold^T \ones_N=1$.
 $S^d\subset\Real^{d\times d}$ denotes open cone of symmetric positive definite matrices.
 The symbol $\Ncal(\cdot\vert \mu, \Sigma)$ denotes a multivariate $d$-dimensional Gaussian density 
 with mean $\mu\in\Real^d$ and covariance $\Sigma\in S^d$:
\begin{equation}
    \mathcal{N}(x\vert \mu, \Sigma) = \frac{1}{(2\pi)^{d/2} \vert\Sigma\vert^{1/2}} e^{-\frac{1}{2}(x-\mu)^T \Sigma^{-1} (x - \mu)}
\end{equation}

\section{THE GAUSSIAN MIXTURE REDUCTION PROBLEM}

A Gaussian Mixture (GM) is a convex combination of $N$ $d$-dimensional Gaussian densities of the form:
\begin{equation} \label{eq:GMM}
    p(x\vert \Theta) = \wbold^T \Ncalbold(x|\mubold,\Sigbold) = \sum_{i=1}^N w_i \mathcal{N}(x\vert \mu_i, \Sigma_i)
\end{equation}
where $\Ncalbold(\cdot|\mubold,\Sigbold)=\begin{bmatrix}
\mathcal{N}(\cdot\vert \mu_1, \Sigma_1) &
\cdots & \mathcal{N}(\cdot\vert \mu_N, \Sigma_N)
\end{bmatrix}^T$, 
$(\mu_i,\Sigma_i) \in \Real^d\times S^d$, and
$\wbold \in\Delta^{N-1}$
is a vector of non-negative weights that sum up to 1 ($\wbold^T \ones_N = 1$).
The set $\Theta = \{\wbold,\mubold,\Sigbold\}\in 
\mathcal{H}_N=\big(\Delta^{N-1}\times\Real^d\times S^d\big)^N$
collects all parameters of the mixture.
The number $N$ of Gaussian components is the {\it size} of the mixture. 
Note that a given value of $\Theta\in \mathcal{H}_N$ uniquely identifies a GM, but different values of $\Theta$ can 
identify the same GM. 
Indeed, considering that any permutation of the parameters in $\Theta$ yields the same mixture, then there are at least $N!$ equivalent parametrizations of a given GM.\footnote{
\noindent
If in a given parameter set $\Theta$ there are two indexes $i$ and $j$, with $i\not = j$, such that 
$\mu_i=\mu_j$ and $\Sigma_i=\Sigma_j$, then there exist infinite parametrizations equivalent to $\Theta$
($w_i$ and $w_j$ can be replaced by any nonnegative pair $\what_i$ and $\what_j$ such that 
$\what_i+\what_j=w_i+w_j$).
Note that in this case the size $N$ of the GM can be reduced to $N-1$, by setting $\what_i=w_i+w_j$ and $\what_j=0$. }  

Roughly speaking, the reduction of a given GM $f(x\vert \Theta^h)$
of size $N^h$ (the superscript $h$ stands for {\it hypothesis} mixture) 
is the process of constructing a GM $g(x\vert \Theta^r)$ with lower size $N^r$ ({\it reduced} mixture) that is similar, in some sense, to the original one.
The symbol $D(\cdot \Vert \cdot)$ will denote a generic {\it dissimilarity measure} 
({\it D-measure}, for short, or {\it deviation}) between two $d$-dimensional distributions, so that
$D(f^h\Vert g^r)$, where $f^h$ and $g^r$ are shorthands for $f(x\vert \Theta^h)$ and $g(x\vert \Theta^r)$, respectively, denotes the dissimilarity between the original and reduced GMs.
In general, D-measures must satisfy the two following requirements, for all pairs of distributions $p$ and $q$:
\begin{align}
& D(p\Vert q)\ge 0,\quad & & \text{\small (nonnegativity)}; \label{eq:Dnonneg}\\  
& D(p\Vert q) = 0\ \Longleftrightarrow\ p=q, & & \text{\small (identity of indiscernibles)}.\label{eq:Dident}
\end{align}
If, for any three distributions, $p$, $q$, $h$, also the following are satisfied
\begin{align}
& D(p\Vert q)=D(q\Vert p),\quad & & \text{\small (symmetry)}; \label{eq:Dsymm}\\ 
& D(p\Vert q)\le D(p\Vert h)+D(h\Vert q),& &\text{\small (triangle inequality)}, \label{eq:Dtriang}
\end{align}
then the dissimilarity $D(\cdot\Vert\cdot)$ is a {\it distance} and defines a metric in the space of distributions.

The Gaussian Mixture Reduction (GMR) problem with respect to a given D-measure is the problem to find the GM $g(x\vert \Theta^r)$, of a given size $N^r$, that minimizes the dissimilarity from the hypothesis GM $f(x\vert \Theta^h)$, with
$N^r< N^h$.
The solution is formally given by
\begin{equation} \label{eq:GMRformulaz}
    {\Theta^r}^* = \argmin_{\Theta^r\in\mathcal{H}_{N^r}} D(f(x\vert \Theta^h)\Vert g(x\vert \Theta^r)).
\end{equation}
This is a complex, non-convex nonlinear constrained optimization problem, which in general does not admit a closed form solution. 
In addition to the presence of (a lot of) local minima, there are at least $N^r!$ global minima.
This happens because, as discussed below equation \eqref{eq:GMM}, any permutation of the parameters of a solution ${\Theta^r}^\ast$ is an equivalent solution.
Note also that any parameter permutation of a local minimum is still a local minimum.
Another issue with the optimization problem \eqref{eq:GMRformulaz} is that few D-measures can be computed in closed form for Gaussian Mixtures. 
Indeed, the evaluation of most D-measures requires a volume integration over the whole domain $\Real^d$.

\section{Dissimilarity Measures}

In this section some D-measures that are used in most existing GMR algorithms are introduced.

\subsection{Kullback-Leibler Divergence}

Given two continuous distributions, $p$ and $q$ on $\Real^d$, the \textit{Kullback-Leibler Divergence} (KLD) of $q$ from $p$ 
(also known as \textit{differential relative entropy}) is defined as \cite{KLD}
\begin{equation} \label{eq:KLdiv}
\DKL(p\Vert q) = \int p(x) \log \frac{p(x)}{q(x)}dx,
\end{equation}
where the integral is over $\textit{supp}(p)$ (the support of $p$), so that
the KLD is well defined when $\textit{supp}(p)\subseteq \textit{supp}(q)$.
When considering GMs, the support is $\Real^d$ and the KL-divergence is always well-defined.
The KLD measures the information loss when $p$ is approximated by $q$ and is a D-measure, because it satisfies  \eqref{eq:Dnonneg} and \eqref{eq:Dident}, but is not a distance, since \eqref{eq:Dsymm} and \eqref{eq:Dtriang} are not satisfied.

The KLD of two Gaussians $\Ncal(\cdot|\mu_i,\Sigma_i)$ and $\Ncal(\cdot|\mu_j,\Sigma_j)$ admits the following closed form 

\vspace{-0.2cm}
\begin{small}
\begin{equation} \label{eq:KLtwoG}
\begin{split}
& \DKL(\Ncal_i\Vert \Ncal_j) =\\
&\frac{1}{2}\big(\trace(\Sigma_j^{-1} \Sigma_i) + \log \frac{\vert \Sigma_j\vert}{\vert \Sigma_i\vert} 
+(\mu_j - \mu_i)^T \Sigma_j^{-1}(\mu_j - \mu_i) - d \big).
\end{split}
\end{equation}
\end{small}
When applied to GMs, the first argument of the KLD is the original GM while the second argument is the reduced one:
\begin{equation} \label{eq:KLdivGM}
\DKL(f(\cdot|\Theta^h)\Vert g(\cdot|\Theta^r)) = 
    \int f(\cdot|\Theta^h) \log \frac{f(\cdot|\Theta^h)}{g(\cdot|\Theta^r)}dx,
\end{equation}
The main issue with the KLD \eqref{eq:KLdivGM} is that it does not have a closed form, 
neither its gradient w.r.t.\ the parameter set $\Theta^r$, useful in numerical minimization, 
has  a closed form. 
One way to compute the KLD between two GMs is through numerical integration of \eqref{eq:KLdivGM} over the whole domain $\Real^d$, a computationally demanding procedure, especially when $d$ is large. 
Also \textit{MonteCarlo} integration methods turn out to be computationally intensive
\cite{Hershey}.
%

A positive feature of the KLD, that will be discussed further on in this paper, is the simplicity of the computation of the barycenter of a GM (see Sec.~\ref{sec:bary}).

\subsection{The ISE dissimilarity}

Another widely used D-measure between two distributions $p$ and $q$ is the Integral Square Error (ISE), called also Integral Square Difference \cite{Scott}, defined as the square of the $L_2$-error
\begin{equation} \label{eq:DISE}
\DISE(p\Vert q) = \int \big(p(x)-q(x)\big)^2 dx =\Vert p - q \Vert_2^2.
\end{equation}
(The name ISE is used when $q$ is aimed at approximating $p$, so that the {\it difference} $p-q$ is in fact an {\it error}.) 
In addition to satisfying the properties \eqref{eq:Dnonneg}-\eqref{eq:Dident}, like the KLD, 
the ISE satisfies also the symmetry \eqref{eq:Dsymm}, but fails to satisfy the triangular inequality \eqref{eq:Dtriang},
and therefore is not a true distance between $p$ and $q$ (note that the square root of $\DISE(p\Vert q)$, which is the  
$L_2$-error norm $\Vert p - q \Vert_2$, satisfies all properties \eqref{eq:Dnonneg}--\eqref{eq:Dtriang}, and therefore is a true distance and defines a metric in the space of distributions).
When applied to GMs $f^h$ and $g^r$ the ISE has a closed form:

\vspace{-0.4cm}
\begin{small}
\begin{align}
&\DISE(f^h\Vert g^r) = \int \big(f(x\vert \Theta^h) - g(x\vert \Theta^r)\big)^2 dx  \notag \\
&= \int f(x\vert \Theta^h)^2 dx - 2\int f(x\vert \Theta^h) g(x \vert \Theta^r) dx + \int g(x\vert \Theta^r)^2 dx  \notag \\
&= J^{hh}(\Theta^h) - 2J^{hr}(\Theta^h,\Theta^r) + J^{rr}(\Theta^r). \label{eq:DISEGMa}
\end{align}
\end{small}

\vspace{-0.4cm}\noindent
$J^{hh}$ and $J^{rr}$ are called \textit{self-likenesses} of $f^h$ and $g^r$, respectively. 
$J^{hr}$ is called \textit{cross-likeness} between $f^h$ and $g^r$.
From expressions $f^h=(\wbold^h)^T\Ncalbold^h$ and $g^r=(\wbold^r)^T\Ncalbold^r$ we get
\begin{equation}
    J^{hr}={\wbold^h}^T \int \Ncalbold^h {\Ncalbold^r}^T dx\, \wbold^r 
    = {\wbold^h}^T H^{hr}\wbold^r.
\end{equation}
Similar expressions hold for $J^{hh}$ and $J^{rr}$
\begin{equation}
    J^{hh}= {\wbold^h}^T H^{hh}\wbold^h,\quad  J^{rr}= {\wbold^r}^T H^{rr}\wbold^r, 
\end{equation}

\vspace{-0.4cm}\noindent
\begin{small}
\begin{equation} \label{eq:Hhhrr}
\hspace{-0.2 cm}\text{where} \  H^{hh}=\int \Ncalbold^h {\Ncalbold^h}^T dx,\quad
H^{rr}=\int \Ncalbold^r {\Ncalbold^r}^T dx.\
\end{equation}
\end{small}

\vspace{-0.4cm}\noindent
Thus, the ISE \eqref{eq:DISEGMa} can be written as
\begin{small}
\begin{equation} \label{eq:DISEGMb}
\!\!\DISE(f^h\Vert g^r) = {\wbold^h}^T H^{hh}\wbold^h - 2 {\wbold^h}^T H^{hr}\wbold^r + {\wbold^r}^T H^{rr} \wbold. 
\end{equation}
\end{small}

\vspace{-0.3cm}\noindent
Of course, $H^{hh}=H^{hh}(\Theta^h)$,  $H^{rr}=H^{rr}(\Theta^r)$ and $H^{hr}=H^{hr}(\Theta^h,\Theta^r)$. 
Exploiting the following identity \cite{WilliamsTh03,Williams06}:

\vspace{-0.3cm}
\begin{small}
\begin{equation}
    \int \mathcal{N}(x\vert \mu_i,\Sigma_i) \mathcal{N}(x\vert \mu_j, \Sigma_j) dx = \mathcal{N}(\mu_i\vert \mu_j, \Sigma_i + \Sigma_j)
\end{equation}
\end{small}

\vspace{-0.3cm}\noindent
the components of $H^{hh}$, $H^{hr}$, $H^{rr}$ can be easily computed
\begin{small}
\begin{equation} \label{eq:Hhhrrij}
\begin{aligned}
\! [H^{hh}]_{ij} & = \mathcal{N}(\mu_i^h\vert \mu_j^h, \Sigma_i^h + \Sigma_j^h), \\
\![H^{rr}]_{kl} & = \mathcal{N}(\mu_k^r\vert \mu_l^r, \Sigma_k^r + \Sigma_l^r), \\
\![H^{hr}]_{ik} & = \mathcal{N}(\mu_i^h\vert \mu_k^r, \Sigma_i^h + \Sigma_k^r), 
\end{aligned}\quad
\begin{aligned} 
i,j & =1,\dots,N^h,\\ k,l &=1,\dots,N^r,
\end{aligned}
\end{equation}
\end{small}

\noindent
Thus, the computation of the ISE between two GMs is an easy and straightforward task. 
In the \cite{WilliamsTh03,Williams06} it is shown that also the computation of the gradient of the ISE with respect to the parameters in $\Theta^r$, useful when pursuing the numerical solution of \eqref{eq:GMRformulaz}, is straightforward.
The availability of closed forms for the computation of the ISE and of its gradient is one of the main advantage over the KLD, for which analogous closed forms do not exist.

\subsection{The NISE dissimilarity}

A dissimilarity measure closely related with the ISE is its normalized version, denoted NISE, defined as

\vspace{-0.4cm}
\begin{small}
\begin{equation}\label{eq:NISEa}
    \DNISE(p\Vert q) = \frac{\int (p(x)-q(x))^2 dx}{\int \big(p^2(x) + q^2(x)\big) dx} =
    \frac{\Vert p - q \Vert_2^2}{\Vert p\Vert_2^2 + \Vert q\Vert_2^2}\in [0,1].
\end{equation}
\end{small}

\vspace{-0.4cm}\noindent
When applied to the GMs $f^h$ and $g^r$, considering the expression \eqref{eq:DISEGMa} for the ISE, we have

\begin{small}
\begin{equation}\label{eq:NISE}
 \DNISE(f^h\Vert g^r) = \frac{J^{hh} - 2J^{hr} + J^{rr}}{J^{hh} + J^{rr}} = 1 - \frac{2J^{hr}}{J^{hh} + J^{rr}} \in[0,1).
\end{equation}
\end{small}

\vspace{-0.4cm}\noindent
Like the ISE, the NISE satisfies the properties \eqref{eq:Dnonneg}--\eqref{eq:Dsymm}, and not 
\eqref{eq:Dtriang}.
The property of the NISE of being confined in the interval $[0,1]$ makes it attractive, in many contexts, like in target tracking, as a measure to evaluate the accuracy of an approximation \cite{Crouse,GMRC}.
Like the ISE, also the gradient of the NISE can be computed in closed form, because the expressions
\eqref{eq:DISE} and \eqref{eq:NISE} are made of the same building blocks $J^{hh}$, $J^{hr}$ and $J^{rr}$.

\subsection{Other dissimilarity measures} \label{sec:OtherD}

Many other D-measures have been considered in the literature for addressing the GMR problem, but are not considered here due to space reasons.
Among these, we mention the Bhattacharyya and the Kolmogorov variational distances
\cite{WilliamsTh03, Williams06}, the Cauchy-Schwarz distance \cite{CauchySchwarz}, the 
Wasserstein distances \cite{AssaPlataniotis,OTGM,Delon} and the class of Composite Transportation Distances (CTD)
\cite{OTGM,CTDGMRA}. Among these, only the Cauchy-Schwarz admits a closed form the evaluation of the cost and of its gradient, while the CTDs can be easily computed solving a Linear Programming problem.

\section{General Structure of GMR algorithms}

Many algorithms have been proposed in the literature with the aim of providing approximate solutions of the GM reduction problem with low computational complexity, so that they could be used in real-time applications.
Most of them make use of two basic steps: {\it greedy reduction} and {\it refinement}
\cite{Crouse}.

\subsection{Greedy Reduction}

This step consists in performing an iterative reduction of the original GM, 
until a GM with the prescribed number of components is achieved with some degree of similarity with the original one, and, hopefully, not far from the optimal solution of \eqref{eq:GMRformulaz}.
Thus, the reduced GM can serve as a good starting point for subsequent refinements.
The basic operations involved at each iteration are \textit{pruning} and \textit{merging} of GM components.

\subsubsection{Pruning}
is the process of removing one or more components from a given Gaussian Mixture according to some criterion. 
Common approaches consist in eliminating components according to criteria based on:
\begin{itemize}
    \item magnitudes of the weights $\wbold$ (thresholding, $k$ smallest, out of prescribed percentile,...),
    \item cost in terms of increment of the chosen D-measure.
\end{itemize}
Pruning according to weights-based criteria is typical in target-tracking contexts \cite{Blackman}. 
The GM components to be removed can be those whose weights are below a fixed or adaptive threshold, or the components associated to the $k$ smallest weights, with $k$ fixed or varying at each step,
or those associated to the smallest weights whose sum is below a prescribed value (typically 0.05). 

This approach is faster than using cost-based criteria, so it can be used to speed up the reduction phase. 
On the other hand, it can lead to a significant worsening of the final approximation, especially in some critical situations%
\footnote{
A critical situation for pruning is the case of a mixture where a group of very similar Gaussian components is associated to very small weights, below the pruning threshold. 
In this case, the merging of the group into a single Gaussian would be better than pruning the components of the group.}.

Cost-based pruning criteria aim at removing from a GM those components whose elimination would give the least increment of the chosen D-measure. 
Thus, for a GM of size $N$ the dissimilarity measure must be computed $N$ times (one computation for each elimination hypothesis), and these computations must be repeated at each pruning step, until the desired number of components $N^r$ is reached.
It is clear that pruning according to cost-based criteria is usually slower than 
pruning according to weights-based criteria, but for some dissimilarity measures it can consistently improve the result.
However, it is important to note that applicability of cost-based criteria is limited, because many dissimilarity measures, like KLD, cannot be computed in closed form for Gaussian mixtures. 
In such cases, approximations \cite{Hershey} or upper bounds \cite{Runnalls} of the dissimilarity measures can be used, at the price of worsening the final result. \\
Whatever the criterion used, after each pruning step the weights must be normalized, so that they sum up to one.

\subsubsection{Merging}
is the process of replacing a specific subset of Gaussian components (a {\it sub-mixture})
of a given GM with only one Gaussian component, according to some criterion.
For a given GM $p(x\vert \Theta)$ of order $N$, we use $p(x\vert \Thetabar)$ to denote a sub-mixture of order $\bar{N}<N$, where $\Thetabar$ denotes a subset of the parameter set $\Theta$ with normalized weights $\wboldbar\in\Delta^{\bar{N}-1}$, while $p(x\vert \Thetabar_u)$ denotes its unnormalized version, i.e.\ the sub-mixture whose weights $\wboldtilde$ are a subset of the weights $\wbold$ of the original mixture, so that $\wboldtilde^T\ones_{\Nbar}< 1$. Of course, $\wboldbar=\wboldtilde/(\wboldtilde^T\ones_{\Nbar})$. 
   Using this notation, the problem of merging the $\Nbar$ components of the sub-mixture $p(x\vert \Thetabar)$
can be formulated similarly to \eqref{eq:GMRformulaz}, 
as the problem of finding the Gaussian $\Ncal(\cdot|\mu,\Sigma)$ with minimal dissimilarity from $p(x\vert \Thetabar)$:
\begin{equation} \label{eq:BSGAdef}
(\mu^\ast,\Sigma^\ast)= \argmin_{(\mu,\Sigma)\in \Real^d\times S^d}
                        D\big(p(\cdot\vert \Thetabar)\Vert \Ncal(\cdot |\mu,\Sigma) \big).
\end{equation}
$\Ncal(\cdot|\mu^\ast,\Sigma^\ast)$ is the \emph{Best Single Gaussian Approximation} (BSGA) of the sub-mixture $p(\cdot\vert \Thetabar)$ for the given $D(\cdot\Vert \cdot)$.
The BSGA can be used to replace in $p(\cdot\vert \Theta)$ the components of the sub-mixture $p(\cdot\vert \Thetabar_u)$, yielding a reduced GM of size $N-\bar{N}+1$. 
The weight assigned to the new component $\Ncal(\cdot|\mu^\ast,\Sigma^\ast)$ is the sum of the weights of the merged components, $\wboldtilde^T\ones_{\Nbar}$, so that the sum of the weights of the resulting mixture remains one.

As a matter of fact, in the literature the merging action has not the interpretation the we have given here.
In most papers, merging consists in replacing a sub-mixture with its {\it barycenter}, defined
as the Gaussian density that minimizes a weighted sum of pairwise dissimilarities \cite{Snoussi}. 
Stated in formulas, the parameters of the barycenter of a sub-mixture
$p(\cdot\vert \Thetabar)=\sum_{i=1}^{\bar{N}} \wbar_i \Ncal(\cdot |\mubar_i,\Sigmabar_i)$ are computed as
\begin{equation} \label{eq:barycenterdef}
(\muhat,\Sigmahat)= \argmin_{(\mu,\Sigma)\in \Real^d\times S^d}
                    \sum_{i=1}^{\bar{N}} \wbar_i D\big(\Ncal(\cdot |\mubar_i,\Sigmabar_i)\Vert \Ncal(\cdot |\mu,\Sigma)\big).
\end{equation}
(the barycenter can equivalently be defined considering in \eqref{eq:barycenterdef}
the unnormalized sub-mixture $p(\cdot\vert \Thetabar_u)$.)
In general, for a given D-measure, the barycenter and the BSGA of a sub-mixture do not coincide.
Nonetheless, the KLD-barycenter and KLD-BSGA coincide \cite{Runnalls}, and it is not difficult to prove that
also the ISE-barycenter and ISE-BSGA coincide, while this is not true for the NISE.
However, only the KLD barycenter/BSGA admits a closed form solution (see Sect.~\ref{sec:bary}).

The choice of which components to merge in a mixture can be made using {\it local} or {\it global} cost criteria.
A local criterion consists in checking, at each step, the dissimilarities between all pairs of components of the given GM, and then merging the pairs with the least dissimilarity (a variant of this approach is to create  {\it clusters} of more than two similar components, and perform the merging of the least dissimilar cluster).
A global criterion consists in checking the effects of all possible merging of pairs of components on the dissimilarity with the original GM, and then choosing to merge the components which yield the least dissimilarity. 

\subsubsection{Combining pruning and merging}
at  each  step  of an iterative  greedy  reduction  algorithm either pruning or merging is applied, the choice depending on criteria involving, in general, both the weights and the evaluation of the dissimilarity between the GMs before and after the reduction, as in \cite{WilliamsTh03}.

\subsection{Refinement}
The greedy reduction step can provide a suitable starting point for a subsequent tuning of the parameters of the reduced mixture.
Most refinements in the literature can be roughly divided in the following three classes:
\begin{itemize}
    \item optimization (e.g.\ \cite{WilliamsTh03,Williams06});
    \item clustering (e.g.\ \cite{AssaPlataniotis,GMRC});
    \item optimization-clustering (e.g.\ \cite{DPHEM,CTDGMRA}).
\end{itemize}
Refinement procedures in the first class consist in using standard numerical methods for solving the optimization problem
\eqref{eq:GMRformulaz}. 
The ISE and NISE are suitable for this approach, because closed forms exist for their computation, together with their gradients, while the KLD is not.
Of course, if a descent method is applied, in general the refinement procedure will provide a local minimum close to the starting point. For this reason, it is extremely important that the greedy reduction provides a good starting point for the refinement.

When the adopted D-measure admits a closed form only between pairs of Gaussians, like the KLD and the 2-Wasserstein, then clustering can be used to decrease the dissimilarity between the original and reduced GMs. In some cases \cite{DPHEM,CTDGMRA}, clustering and optimization are combined together.

\subsection{Consistency issues in the pipeline of reduction operations}

It is important to stress that, due to the abundance of local minima for the GMR problem \eqref{eq:GMRformulaz}, the outcome of all the mentioned refinement approaches strongly depends on the starting point of the algorithm, which is the reduced GM provided by the greedy reduction step.
For this reason, we believe that it is important that in the pipeline of reduction steps (greedy pruning/merging operations + refinement by clustering/optimization) the same D-measure should be used (consistency issue).
If the initial point of an algorithm aimed at optimizing a given D-measure is provided by a greedy reduction performed on the basis of a different D-measure, the result of the refining phase can be far from optimality.

Indeed, most GMR algorithms proposed in the literature are made of a pipeline of steps in which different D-measures are considered (inconsistency), and a further D-measure is used for evaluating their performances (typically, the NISE is used for final assessment).
The reason is that, depending on the chosen D-measure, some reduction steps may have or not a closed form or can be completed with simple or cumbersome computations.
To the best of our knowledge, the issue of consistency among the steps of a reduction procedure has never been raised before. 
We refer to Sections \ref{sec:summary} and \ref{sec:consist} for a further discussion on this issue. 
\section{KLD and ISE BSGA and Barycenters} \label{sec:bary}

One attractive point of the KLD as a dissimilarity measure is that the BSGA of a sub-mixture and its barycenter coincide \cite{Runnalls}, and can computed in closed form.
Given an unnormalized submixture $p(\cdot|\Thetabar_u)$ of size $\Nbar$ of a given GM $p(\cdot|\Theta)$,
with $\Thetabar_u=\{\wboldtilde,\muboldbar,\Sigboldbar\}\subset \Theta$,
the mean and covariance of the barycenter are computed as follows:

\vspace{-0.4cm}
\begin{small}
\begin{equation} \label{eq:KLDbary}
\begin{split}
    & \mu^\ast =   \frac{1}{\wboldtilde^T\ones_{\Nbar}}\sum_{i=1}^{\bar{N}} \wtilde_i \mubar_i \\
    & \Sigma^\ast= \frac{1}{\wboldtilde^T\ones_{\Nbar}}\sum_{i=1}^{\bar{N}} \wtilde_i
               \big(\Sigmabar_i + (\mubar_i - \mu^\ast)(\mubar_i - \mu^\ast)^T)
\end{split}
\end{equation}
\end{small}

\noindent
In the merging step of a GMR algorithm, the Gaussian component $\Ncal(\cdot|\mu^\ast,\Sigma^\ast)$ replaces the unnormalized sub-mixture $p(\cdot|\Thetabar_u)$ in the GM $p(\cdot|\Theta)$, with assigned weight $w^\ast=\wboldtilde^T\ones_{\Nbar}$.
This replacement is also called \textit{moment-preserving merging}, because it preserves the mean and the covariance of the original GM \cite{Runnalls}.

The ISE barycenter and the ISE BSGA coincide as well, although they can not be computed in a closed form, due to trascendental equations.
Therefore, most GMR algorithms compute the merging actions using KLD-barycenters, through equations \eqref{eq:KLDbary} 
(the KLD barycenter is also frequently used in clustering steps, e.g.\ when updating the centroids in a $k$-means scheme).
However, it is important to stress that the numerical computation of the ISE barycenter is eased by the availability of a closed form of the ISE gradient.

\section{Summary on KLD, ISE and NISE features} \label{sec:summary}

This section lists some features of the KLD, ISE and NISE that explain the reason why most GMR algorithms jump from one D-measure to another in the various steps \cite{Crouse}.

\vspace{0.1cm}
{\small 
\noindent {\it Features of the KLD:}
\begin{itemize}  \setlength\itemsep{0.08cm}
\item[a1)] 
KLD BSGA and KLD barycenter \underline{coincide}\\
({\it moment preserving merging} of sub-mixtures);
\item[a2)] 
closed form \eqref{eq:KLDbary} \underline{available} for the KLD barycenter;
\item[a3)] 
closed form for the KLD and KLD gradient \underline{not available}:\\
-\ numerical computations are needed; \\
-\ closed-form only for KLD of two Gaussians \eqref{eq:KLtwoG};
\item[a4)] pruning operations may dramatically increase the KLD \\
(KLD is prone to merge components rather than cut them);
\end{itemize}
}

\noindent
Although properties (a1),(a2) make attractive the KLD for merging, (a3) makes it unsuitable 
in GMR optimization problems \eqref{eq:GMRformulaz}.
Property (a4) is obvious when looking at \eqref{eq:KLdivGM}: if a rather isolated component of $f^h$ is pruned then there are points where the ratio $f^h(x)/g^r(x)$ in the logarithm of \eqref{eq:KLdivGM} can take on high values
(indeed, the KLD between two Gaussians \eqref{eq:KLtwoG} grows quadratically with the distance between the means).
 
\vspace{0.1cm}
{\small 
\noindent {\it Features of the ISE dissimilarity:}
\begin{itemize}  \setlength\itemsep{0.08cm}
\item[b1)] 
ISE BSGA and ISE barycenter \underline{coincide};
\item[b2)] 
a closed form is \underline{not available} for the ISE barycenter\\
(numerical computation is needed);
\item[b3)] 
closed forms for ISE and ISE gradient are \underline{available};
\item[b4)] in some situations, optimizing the ISE resembles pruning. 
\end{itemize}
}

\noindent
Property (b2) makes the ISE less appealing than KLD for merging, although thanks to (b3), it can be used as a criterion for merging components of a GM.
Indeed, in \cite{WilliamsTh03,Williams06}, during the greedy reduction phase, at each step, the pair of Gaussians with the lowest impact on the ISE between the original and the reduced GMs are merged.
However, the KLD barycenter is used for merging (inconsistency).
In general, property (b3) makes the ISE attractive as a cost function to be numerically optimized, as in \cite{WilliamsTh03,Williams06} during the refinement phase.
The property (b4) will be illustrated in the example section \ref{sec:consist}. 

\vspace{0.1cm}
{\small
Features of the NISE dissimilarity:
\begin{itemize}
\item[c1)] 
NISE BSGA and NISE barycenter \underline{do not coincide}\\
(they are {\it close} when the ISE is close to 0);
\item[c2)] 
closed forms for BGSA and barycenter \underline{not available}
(numerical computations are needed);
\item[c3)] 
closed forms for NISE and NISE gradient are \underline{available};
\item[c4)] in some situations, optimizing the NISE resembles pruning.
\item[c5)] the NISE takes values in $[0,1]$ ($[0,1)$ for GMs); 
\item[c6)] in most cases the NISE tends to yield reduced GMs with lower covariance, w.r.t.\ the ISE.
\end{itemize}
} 

\noindent
The ISE and NISE share the properties (c2),(c3),(c4), and thus the same comments are appropriate.
Like the ISE, merging with NISE can be done by numerically solving the optimization problem \eqref{eq:BSGAdef} or \eqref{eq:barycenterdef}, although obtaining different solutions in the two cases.
Due to property (c5) the NISE is often used to compare results of different reducing algorithms (e.g.\ \cite{Crouse,GMRC}), although it is never used as the only cost function inside the reduction pipeline.
Property (c6) directly comes from the NISE expression \eqref{eq:NISE}, by considering that the value of $J^{rr}$ in the denominator tends to be large when the covariances of the reduced mixture are small.

\section{ Numerical tests on consistency} \label{sec:consist}

\begin{figure}[hbtp]
    \centering
    \includegraphics[scale=0.15]{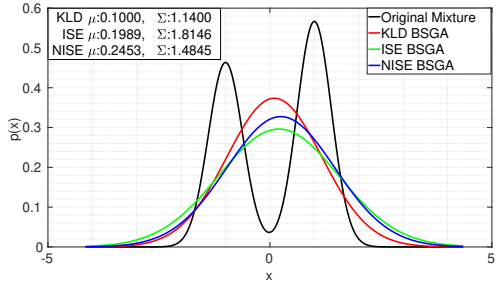}
    \vspace{-0.3cm}
    \caption{BSGA of components with means $\mu^h_1=-1$, $\mu^h_2=1$.}
    \label{fig:1}
\end{figure}

\begin{figure}[hbtp]
    \centering
    \vspace{-0.3cm}
    \includegraphics[scale=0.15]{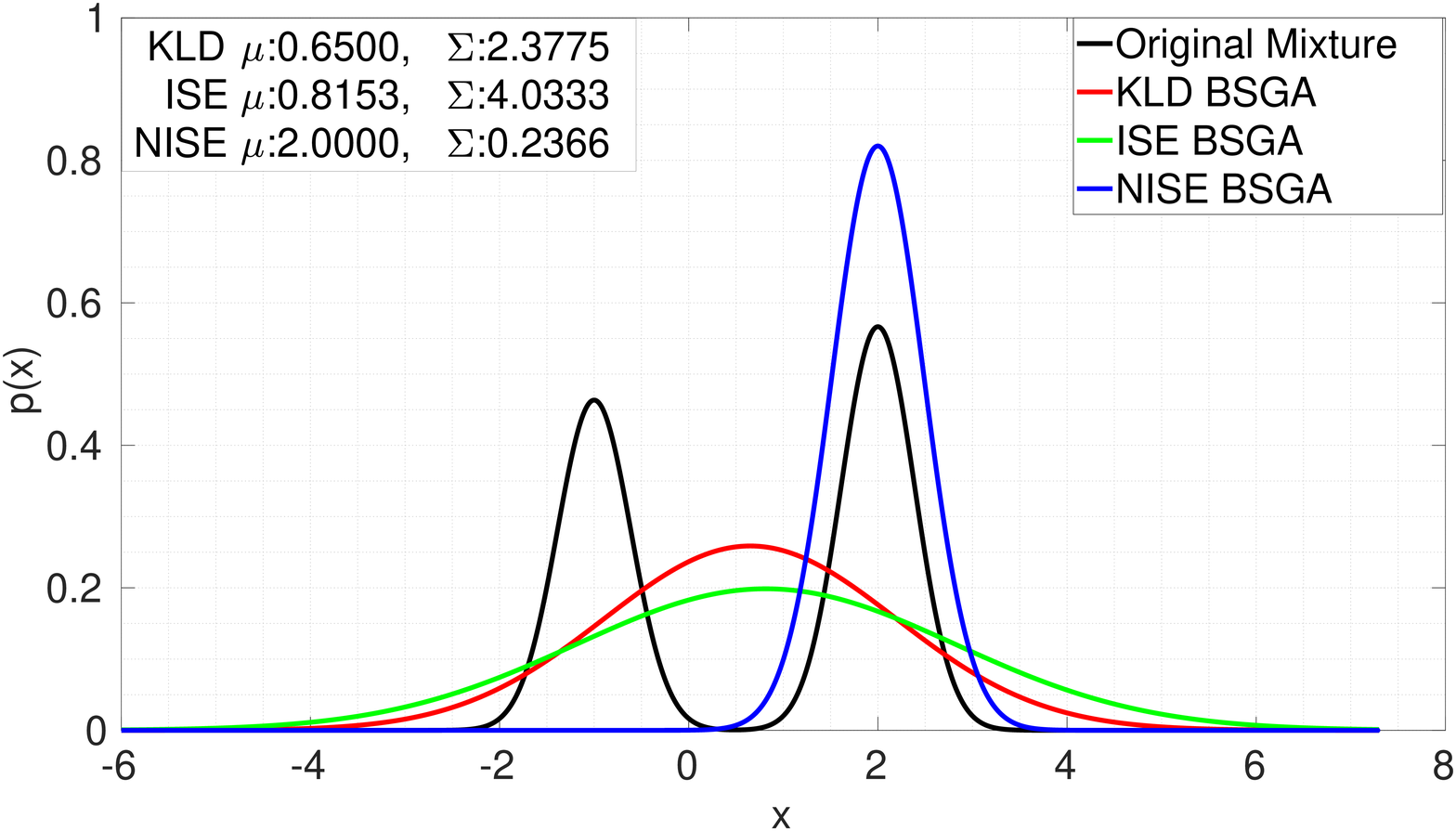}
    \vspace{-0.3cm}
    \caption{BSGA of components with means $\mu^h_1=-1$, $\mu^h_2=2$.}
    \label{fig:2}
\end{figure}

\begin{figure}[hbtp]
    \centering
    \vspace{-0.3cm}
    \includegraphics[scale=0.15]{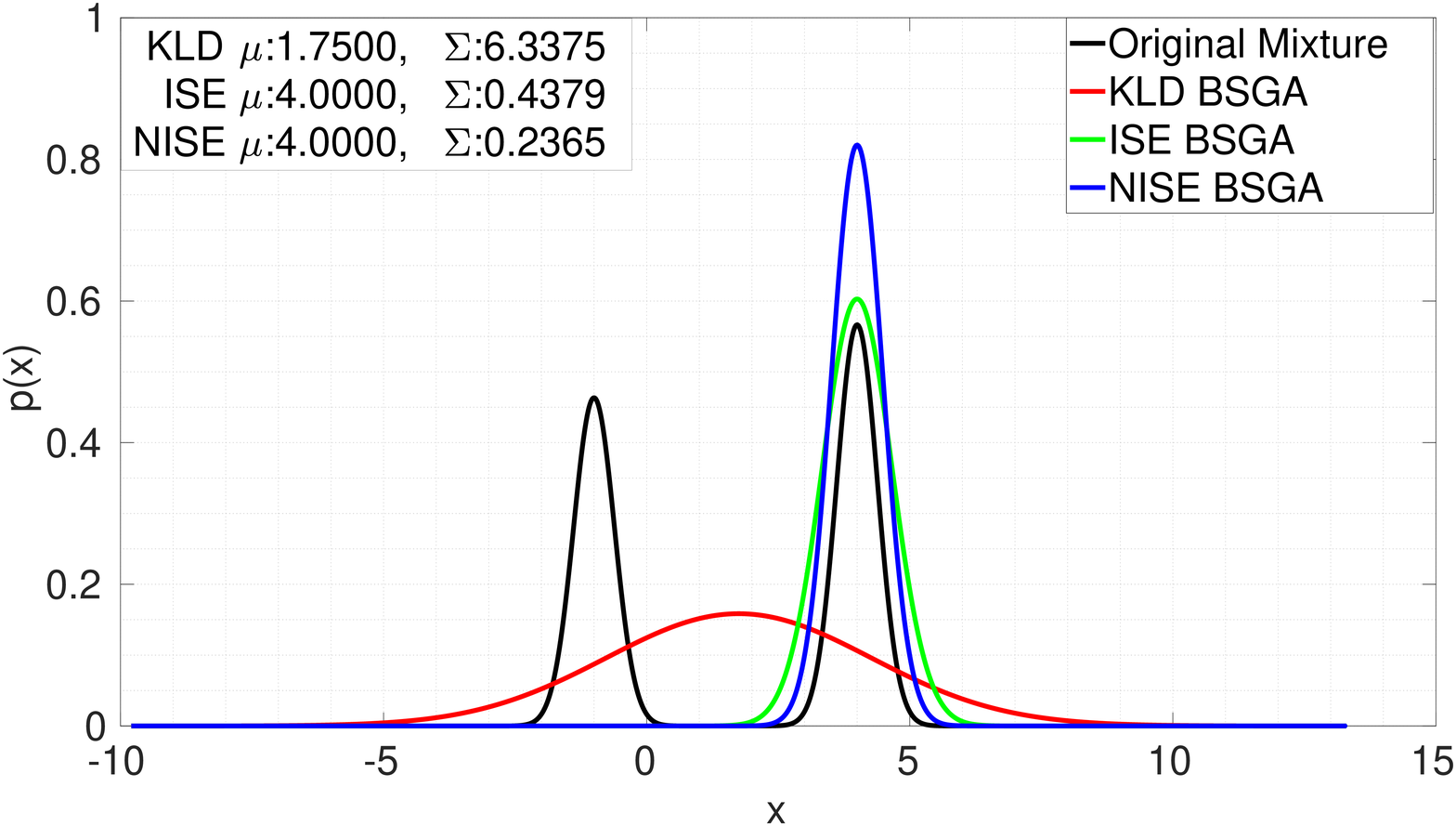}
    \vspace{-0.3cm}
    \caption{BSGA of components with means $\mu^h_1=-1$, $\mu^h_2=4$.}
    \label{fig:3}
\end{figure}

\begin{figure}[hbtp]
    \centering
    \includegraphics[scale=0.15]{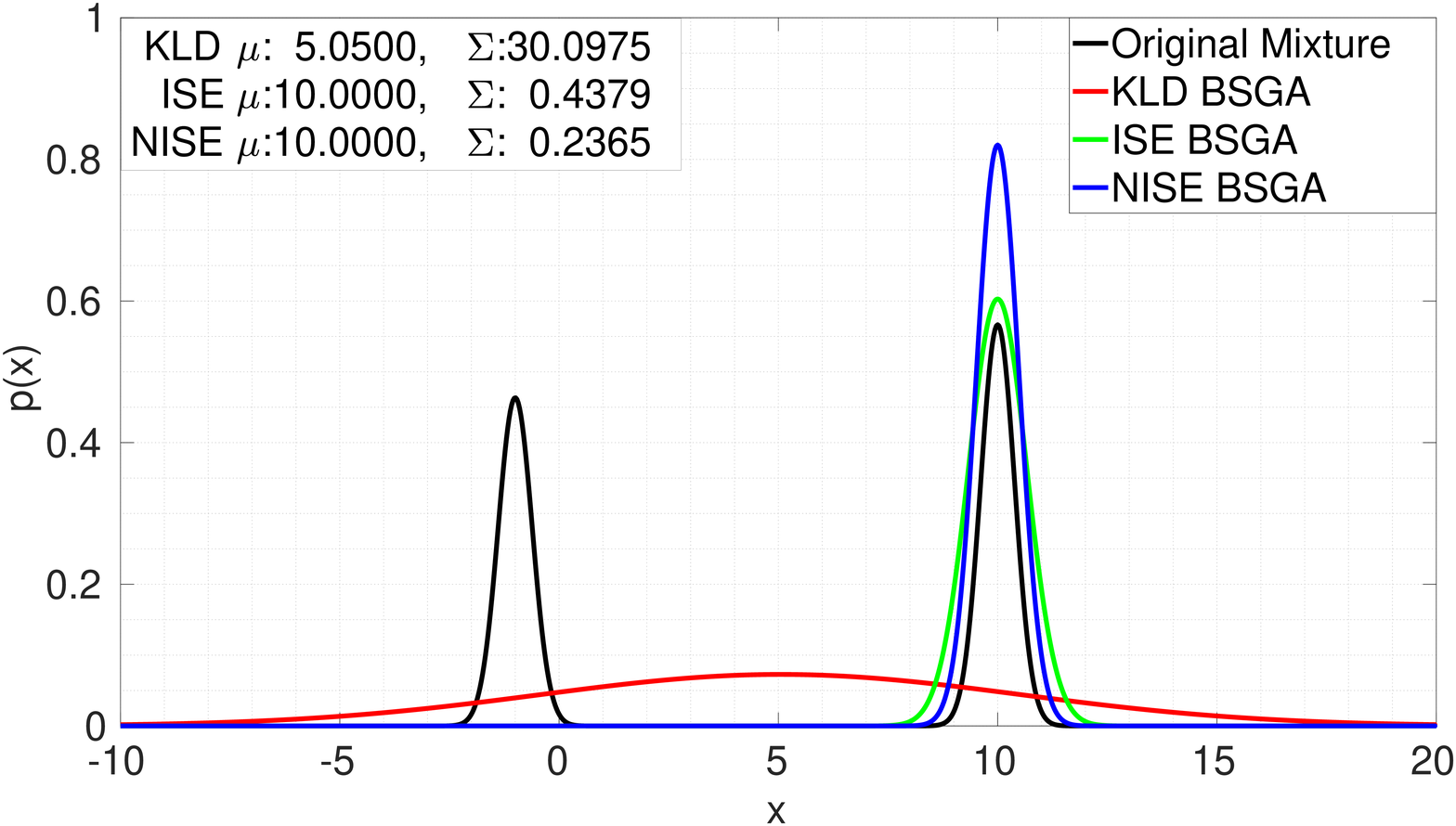}
    \vspace{-0.3cm}
    \caption{BSGA of components with means $\mu^h_1=-1$, $\mu^h_2=10$.}
    \label{fig:4}
\end{figure}

The features and issues listed in the previous section are discussed here on a couple of case studies.
To make things clear, the very simple reduction problem of a GM of size two ($N^h=2$) to a GM of size one ($N^r=1$) is considered.
This is equivalent to the problem of finding the BSGA of a submixture made of two Gaussian components.
Let $(w^h_1,\mu_1^h,\Sigma^h_1)$ and $(w^h_2,\mu_2^h,\Sigma^h_2)$ denote the parameters of the components of the original GM $f^h$.
In conditions of perfect symmetry of the components in $f^h$  
($w^h_1=w^h_2=0.5$, $\Sigma^h_1=\Sigma^h_2$),
the mean of the KLD-barycenter \eqref{eq:KLDbary} is the arithmetic mean
$\mu_\KLD^\ast=(\mu^h_1+ \mu^h_2)/2$.
A rather challenging and enlightening situation is when the original GM is not perfectly symmetric, i.e.\ when
the covariances are the same, but the weights are slightly different.
Below, we compare the KLD, ISE and NISE BSGAs
for a set of four GMs of size $N^h=2$ with parameters
\begin{equation}
\begin{aligned}
&\begin{bmatrix} w_1^h &  w_2^h\end{bmatrix} =\begin{bmatrix} 0.45 & 0.55 \end{bmatrix},\\
&\begin{bmatrix} \Sigma_1^h & \Sigma_2^h \end{bmatrix} =\begin{bmatrix} 0.15 & 0.15 \end{bmatrix},\\
& \mu_1^h=-1, \qquad \mu_2^h\in\{1,2,4,10\}.
\end{aligned}
\end{equation}
Fig.s \ref{fig:1}--\ref{fig:4} report the original GM (black) and the reduced mixtures (BSGAs) according to the respective D-measures.

\begin{figure}[hbtp]
    \centering
    \vspace{-0.5cm}
    \includegraphics[scale=0.17]{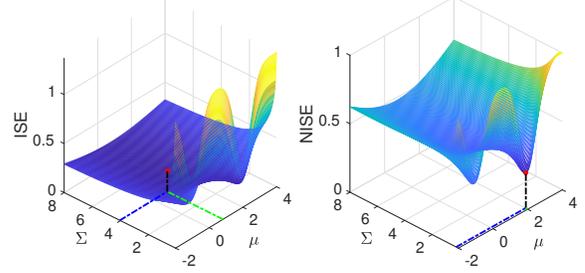}
    \caption{Example of ISE and NISE surfaces for $\mu^h_1=-1$, $\mu^h_2=2$.}
    \label{fig:ISENISESurf}
    \vspace{-0.5cm}
\end{figure}
\vspace{-0.8cm}
\noindent
It is interesting to analyze the behaviors of the three BSGAs when $\mu_2^h$ moves away from $\mu_1^h$, so that the overlapping of the two components of the original GM progressively decreases and the covariance of the GM increases (quadratically with $\mu_2^h-\mu_1^h$, as in \eqref{eq:KLDbary}).
First thing to note is that the mean $\mu_\KLD^\ast$ is always roughly in the middle between $\mu_1^h$ and $\mu_2^h$, since it is the weighted mean of $\mu_1^h$ and $\mu_2^h$ \eqref{eq:KLDbary},
while the means $\mu_\ISE^\ast$ and $\mu_\NISE^\ast$ vary considerably when $\mu_2^h$ moves away from $\mu_1^h$.
Only when $\mu_2^h=1$, close to $\mu_1^h$, the shapes and locations of the three BSGAs are all similar 
(Fig.~\ref{fig:1}). 
For $\mu_2^h\in\{2,4,10\}$ the NISE BSGA is very similar to the Gaussian component of the original GM
with the highest weight. 
Indeed, the mean value is almost the same, $\mu_\NISE^\ast\approx \mu_2^h$, while the covariance
$\Sigma_\NISE^\ast$ is a bit larger than $\Sigma_2^h$.
The ISE BSGA has an intermediate behavior: when $\mu_2^h\in\{1,2\}$, rather close to $\mu_1^h$, the mean 
$\mu_\ISE^\ast$ is close to $\mu_\KLD^\ast$, while the covariance $\Sigma_\ISE^\ast$ is larger than
$\Sigma_\KLD^\ast$ (Fig.s \ref{fig:1}-\ref{fig:2}).
When $\mu_2^h\in\{4,10\}$, rather far from $\mu_1^h$, the ISE BSGA is more similar to the NISE BSGA, and hence to
the Gaussian with the highest weight in the original mixture (Fig.s \ref{fig:3}-\ref{fig:4}).
Indeed, $\mu_\ISE^\ast\approx \mu_\NISE^\ast\approx \mu_2^h$, while the covariance is larger,
$\Sigma_\ISE^\ast > \Sigma_\NISE^\ast$, so that the peak values of the original GM and that of the ISE BSGA are very close. On the contrary, due to the smallest covariance, the peak of the ISE BSGA is considerably higher than the peak of the original GM.
Note that the effect of ISE and NISE optimization, when $\mu_2^h\in\{4,10\}$, is the same as pruning, while the effect of KLD optimization is the same as merging (see the properties (a4),(b4),(c4) in the previous Section). 
It is important to stress that the ISE and NISE BSGAs have been computed by numerically solving 
\eqref{eq:BSGAdef}, due to properties (b2),(c2), and we have found that in general local minima are present.
For instance, the surface plot in Fig.~\ref{fig:ISENISESurf} shows that the NISE has a local minimum with the mean coinciding with $\mu_1^h=-1$, and this is true for all the four cases considered. 
Moreover, we have found that when $\mu_2^h\in\{4,10\}$ the ISE has a local minimum close to the KLD BSGA
($\mutilde_\ISE=2.0419$, $\Sigtilde_\ISE=11.1479$).
Indeed, a descent algorithm that is initialized with the KLD BSGA, which is easy to compute using \eqref{eq:KLDbary}, will converge toward this local minimum, failing to reach the global one.
These considerations should warn from choosing as initial point for the ISE optimization a mixture obtained by a greedy reduction carried out according to KLD-based criteria or KLD merging, as done in 
\cite{WilliamsTh03,Williams06}.
Of course, in real-life applications, where both $N^h$ and $d$ can be large, local minima are extremely difficult, if not impossible, to detect and to avoid. 

The simple reduction examples reported in Fig.s \ref{fig:1}--\ref{fig:4}
clearly show that the solution of the GMR problem as formulated in \eqref{eq:GMRformulaz}
strongly depends on the choice of the D-measure.
In particular, looking at Fig.s \ref{fig:3}--\ref{fig:4} it is apparent that the GMR carried out minimizing the KLD, yields a reduced GM whose shape can be very different from the original one, although preserving the mean and the covariance.
In contrast, the GMR carried out minimizing the ISE yields a reduced GM that rather faithfully overlaps with the most significant portion of the original GM, while pruning the least significant components.
The larger covariance, with respect to that of the corresponding portion of the original GM, is due to the uncertainty added by the reduction process.
In this respect, it is not advisable the use of the NISE in the GMR problem 
 \eqref{eq:GMRformulaz} since it yields reduced GMs with {\it artificially} small covariances (see property (c6)).

To further analyze the issue of consistency, a GMR problem taken from \cite{WilliamsTh03} is considered below, where a one-dimensional GM of size $N^h=5$ has to be reduced to size $N^r=2$.
The parameters of the original GM are:

\vspace{-0.4 cm}
\begin{small}
\begin{equation} \label{eq:testGM}
\begin{split}
    \wbold^h = &\begin{bmatrix}0.083 & 0.167 & 0.25 & 0.333 & 0.167 \end{bmatrix}^T \\ \boldsymbol{\mu}^h = &\begin{bmatrix}\ 1\ & 2 &\ 3 & 4 & 10 \end{bmatrix}^T \\
    \boldsymbol{\Sigma}^h = &\begin{bmatrix}0.1 & 20 & 2 & 2 & 2 \end{bmatrix}^T
\end{split}
\end{equation}
\end{small}

\noindent
Fig.~\ref{fig:williamsKLD} reports the GM obtained by applying the greedy reduction algorithm proposed  by Williams in \cite{Williams06}:
\begin{figure}[hbtp]
    \centering
    \includegraphics[scale=0.19]{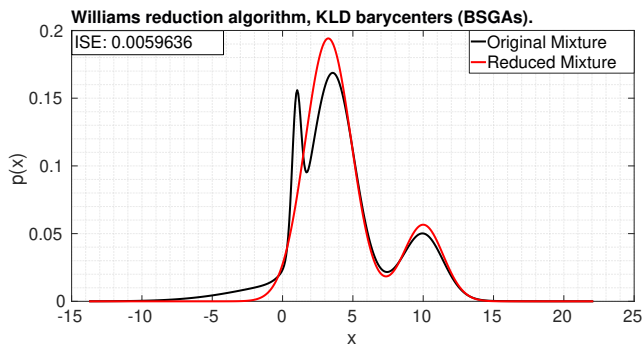}
    \caption{GMR by Williams' greedy reduction and KLD BSGAs.}
    \vspace{-0.3 cm}
    \label{fig:williamsKLD}
\end{figure}

\noindent
In the William's algorithm, pairs of Gaussian components are merged into their KLD-barycenters 
(moment preserving merge), but the selection of the components to be merged or pruned is made according to the ISE cost.
At first, the components 3 and 4 (out of 5) are merged, then the component 2 (out of 4) is pruned, and finally the components 1 and 2 (out of 3) are merged. 
The resulting GM and the original GM have an ISE dissimilarity of 0.0059636.\\
Next we consider a ISE-consistent modification of William's algorithm,
where the ISE-BSGA, computed with the gradient descent, is used instead of the KLD-BSGA (Fig.\ \ref{fig:williamsISE}):

\begin{figure}[hbtp]
    \centering
    \includegraphics[scale=0.19]{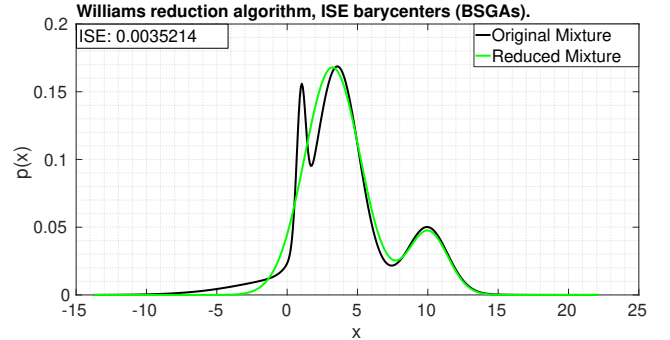}
    \caption{GMR by Williams' greedy reduction and ISE BSGAs.}
    \label{fig:williamsISE}
\end{figure}

\begin{figure}[hbtp]
    \centering
    \includegraphics[scale=0.185]{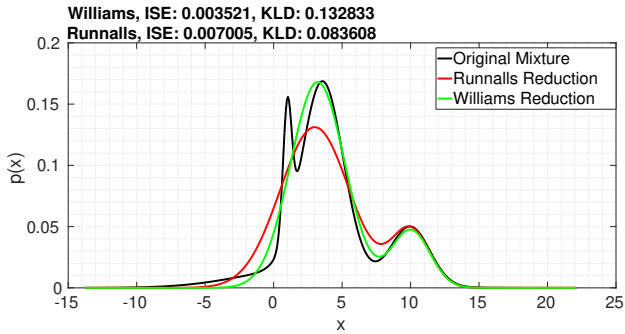}
    \caption{Modified Williams' (ISE-consistent) vs Runnalls' (KLD-consistent) algorithms.}
    \label{fig:williamsvsrunnalls}
\end{figure}

\noindent
Using the ISE-consistent merging a different sequence of reduction steps comes up: the components 3 and 4 (out of 5) are merged first, followed by components 2 and 3 (out of 4), and finally 1 and 2 (out of 3), without any case of pruning. 
  Thus, merging components into the ISE-barycenters has led the algorithm to prefer merging over pruning in the second iteration, unlocking a different reduction solution in the iterative process.
  It can be noticed that the final score is almost halved in magnitude w.r.t.\ to the KLD-barycenter case.
We got similar results applying the modified ISE-consistent William's algorithm to much more complicated situations, not reported here, so that we ascertain that the choice of carrying out a pipeline of actions consistent with a unique D-measure, has a tendency to provide more accurate reduced GMs, according to the chosen D-measure.
\newline
To further support our arguments about consistency, the performances of \textit{fully consistent} algorithms are compared on the same test GM \eqref{eq:testGM}.
As KLD-consistent algorithm we consider the one proposed by Runnalls in \cite{Runnalls},
where an upper bound on the KLD of the GMs before and after the moment-preserving merge has been
proposed and used as a cost criterion for merging.
Runnalls' algorithm does not consider pruning because only a bound on the KLD-cost of merging actions has been developed in \cite{Runnalls}. This is not a real drawback because the KLD naturally does not foster pruning (property (a4)).
By comparing the Runnalls' algorithm with the modified \textit{ISE-consistent} Williams' algorithm, we get the result of Fig.\ \ref{fig:williamsvsrunnalls}.
The KLD values have been computed through numerical integration here, due to (a3).
As expected, by solving the GMR problem in a consistent way, fairly different solutions are found. 
The KLD-consistent reduced GM is such to preserve the mean and the covariance of the original GM, but the shape is rather different from the original one.
On the contrary, the ISE-consistent reduced GM has a shape that quite accurately traces a good portion of the original GM (the components with larger weights), and prunes the remaining (the thin peak on the left).

By looking at the examples of Fig.s \ref{fig:1}--\ref{fig:4} and of Fig.~\ref{fig:williamsvsrunnalls}, we ascertain that in applications aimed at computing \textit{maximum a posteriori (MAP)} estimates, GM reductions according to ISE measure should be preferred (ISE-consistency), while in applications
where \textit{minimum mean square error (MMSE)} estimates are needed, the KLD measure is the right choice.
Nevertheless, the computational burden introduced by using a specific reduction pipeline can be the main hindrance. In general, algorithms based on global dissimilarity measures require more computational power and, if working in real-time scenarios, one might prefer a trade-off between accuracy and efficiency by recurring to lighter solutions; on the other hand, if time is not a constraint, or the available computational power is significant, one should consider GMR pipelines fully consistent  with a unique D-measure of interest.

\section{CONCLUSIONS}

In this work, after the presentation of a general view of the GMR problem and algorithms,  
the features of the most used D-measures (KLD, ISE, NISE), have been reviewed in some detail and illustrated on simple but informative case studies.
In particular the issue of consistency in the pipeline of actions taken
in a GMR algorithm has been discussed.
The analysis and the tests performed reveal that all actions
should be consistent with a single D-measure in order to obtain a reduced GM 
hopefully close to the optimal one.
Unfortunately, the considered D-measures have complementary characteristics, in terms of existence of closed forms and ease of computation, and for this reason most GMR algorithms inconsistently combine their use, thus yielding results that in some situation can be far from optimality.
In a future work we intend to include other D-measures in the analysis.

\end{document}